\documentclass[10pt,conference,compsocconf]{IEEEtran}
\usepackage{times}
\usepackage{graphics} 
\usepackage{epsfig} 
\usepackage{mathptmx} 
\usepackage{amsmath} 
\usepackage{amssymb}  
\usepackage{newtxtext, newtxmath}
\usepackage{caption}
\usepackage{tikz}
\usetikzlibrary{arrows,positioning,patterns,decorations.pathmorphing}
\usepackage{pgfplots}
\usepackage{booktabs}
\usepackage{colortbl}%
  \newcommand{\myrowcolour}{\rowcolor[gray]{0.925}}
\usepackage{pifont}
\usepackage{subcaption}
\usepackage{copyrightbox}

\newcommand{\cmark}{\ding{51}}%
\newcommand{\xmark}{\ding{55}}%

\newcommand{\norm}[1]{\left\lVert#1\right\rVert}

\newcommand{\vect}[1]{\mathbf{#1}}
\newcommand{\image}{\mathcal{I}}

\ifCLASSINFOpdf
\else
\fi
\begin{document}
\pagenumbering{gobble}
%

\title{\textbf{\Large Towards a Unified Approach to Homography Estimation\\[-1.5ex]Using Image Features and Pixel Intensities}\\[0.2ex]}

\author{\IEEEauthorblockN{~\\[-0.4ex]\large Lucas Nogueira, Ely C. de Paiva\\[0.3ex]\normalsize}
\IEEEauthorblockA{School of Mechanical Engineering\\
University of Campinas\\
Campinas, SP, Brazil\\
{\tt [lucas.nogueira]|[ely]@fem.unicamp.br}}
\and
\IEEEauthorblockN{~\\[-0.4ex]\large Geraldo Silveira\\[0.3ex]\normalsize}
\IEEEauthorblockA{Robotics and Computer Vision research group\\
Center for Information Technology Renato Archer\\
Campinas, SP, Brazil \\
{\tt Geraldo.Silveira@cti.gov.br}}
}


%


\maketitle

\begin{abstract}
The homography matrix is a key component in various vision-based robotic tasks. Traditionally, homography estimation algorithms are classified into feature- or intensity-based. The main advantages of the latter are their versatility, accuracy, and robustness to arbitrary illumination changes. On the other hand, they have a smaller domain of convergence than the feature-based solutions. Their combination is hence promising, but existing techniques only apply them sequentially. This paper proposes a new hybrid method that unifies both classes into a single nonlinear optimization procedure, applies the same minimization method, and uses the same homography parametrization and warping function. Experimental validation using a classical testing framework shows that the proposed unified approach has improved convergence properties compared to each individual class. These are also demonstrated in a visual tracking application. As a final contribution, our ready-to-use implementation of the algorithm is made publicly available to the research community.
\end{abstract}



%
\IEEEpeerreviewmaketitle

\section{Introduction}
The homography matrix is a key component in computer vision. It relates corresponding pixel coordinates of a planar object in different images, and has been used in a variety of vision-based applications such as image mosaicing
\cite{faugeras_book2001}, visual servoing \cite{benhimane_07} and object grasping \cite{neuberger2019object}. The homography estimation task can be formulated as an Image Registration (IR) problem. IR can be defined as a search for the parameters that best define the transformation between corresponding pixels in a pair of images. Solutions to this problem involve the definition of at least four important characteristics \cite{brown1992survey}: the information space, the transformations models, the similarity measures, and the search strategy.

With respect to the information space, the vast majority of vision-based algorithms use a Feature-Based (FB) approach. In this class, firstly an extraction algorithm searches each image for geometric primitives and selects the best candidates. Then, a matching algorithm establishes correspondences between features in different images. Afterwards, the actual estimation takes place. However, both the extraction and matching steps are error-prone and can produce outliers that affect the quality of the estimation. Additionally, by using only a sparse set of features, these algorithms may discard useful information.

In contrast, Intensity-Based (IB) methods have no extraction and matching steps. These methods are also referred to as direct methods since they exploit the pixel intensity values directly. This allows the estimation algorithm to work with more information than FB methods and does not depend on particular primitives. Thus, it leads to more accurate estimates and is highly versatile. However, an important drawback is that they require a small interframe displacement, i.e., a sufficient overlapping between consecutive images.

The algorithms presented in this work use multidimensional optimization methods as the main search strategy for the image registration problem. When formulated as such, an initial solution is iteratively refined using a nonlinear optimization method. Specifically, the algorithms presented here are derived from the Efficient Second-order Minimization method (ESM) \cite{benhimane2004real}. Its advantages include both a higher convergence rate and a larger convergence domain than standard iterative methods. It allows for a second-order approximation of the Taylor series without computationally expensive Hessian calculations.

The use of the ESM framework has shown remarkable results for IB methods. However, its application within FB methods has been limited so far. As discussed, the two classes of estimation methods have complementary strengths. This work aims to develop a hybrid method that exploits their advantages and reduces their shortcomings. The proposed algorithm is made available as ready-to-use ROS \cite{quigley2009ros} packages and as a C++ library. In particular, a homography-based visual tracking application is also developed. In summary, our contribution is the development of a vision-based algorithm that:
    \begin{itemize}
        \item unifies the intensity- and feature-based approaches to homography estimation into a single nonlinear optimization problem;
        \item solves that problem using the same efficient minimization method, homography parametrization, and warping function;
        \item can be applied in real-time settings, such as for homography-based visual tracking as experimentally demonstrated in this paper; and
        \item its ready-to-use implementation is made publicly available for research purposes as a C++ library and as a ROS package.
    \end{itemize}
    
The remainder of this article is organized as follows. Section~\ref{related} presents the related works, whereas Section~\ref{sec:method} describes the proposed unified approach.  Section~\ref{sec:results} then reports the benchmarking experiments and the application of the proposed algorithm to visual tracking. Finally, the conclusions are drawn in Section~\ref{sec:conc}, and some references are given for further details.
    
\section{Related Works}
\label{related}

The main distinction between IB and FB methods regards their information space. Indeed, on one hand FB requires the extraction and association of geometric primitives in different images before the actual estimation can occur. These primitives can be, e.g., points and lines \cite{faugeras_book2001}\cite{szeliski2007image}. IB methods simultaneously solves for the estimation problem and pixel correspondences with no intermediate steps \cite{anadan_direct}\cite{silveira_phdthesis}.

The transformation model dictates which parameters are estimated. For example, the original Lucas-Kanade \cite{lucas1981} algorithm only estimated translations in the image space. This was later extended to more sophiscated warp functions \cite{bergen1992hierarchical}. Simultaneous Localization And Mapping (SLAM) algorithms commonly use IR to perform the pose and structure estimation \cite{zhang2015visual}. The homography matrix is often used as a transformation model when dealing with predominantly planar regions of interest \cite{silveira2008efficient}\cite{orbslam2015}\cite{detone2016}. Illumination parameters may also be considered as a component of the transformation model, e.g., in \cite{silveira_jmiv_oillum}.

The quality of the IR is defined by a similarity measure. When an optimization method is applied, this measure is often used as a cost function, such as the Sum of Squa\-red Differences (SSD) \cite{lucas1981}\cite{silveira_ijcv10}. Other possibilities include correlation-based metrics \cite{evangelidis2008parametric}\cite{fonseca1996registration} and mutual information \cite{viola1997alignment}.

The last component of IR algorithms is the search strategy. Most real-time applications use a multidimensional optimization approach based on gradient descent. They use the first and second derivatives of the similarity measures with respect to the transformation parameters. The ESM algorithm is such an example, and is applied in the proposed method. Alternative optimization approaches include Gauss-Newton and Levenberg-Marquardt \cite{baker2004}. All of these techniques are most suited to applications with small interframe displacements. Indeed, global techniques are too computationally expensive to be applied in real-time settings. A more thorough review and comparison of image registration algorithms can be found in \cite{Zitova03imageregistration}\cite{mastersthesis}.

As for the existing techniques that combine IB and FB methods, their overwhelming majority only applies them sequentially, e.g., \cite{jianchao2001image}\cite{ladikos2007real}. In sequential strategies, a FB technique is firstly considered and then its estimated parameters are fed as the initial guess to some IB optimization. This standard combination scheme is thus not optimal and is more time consuming. An exception to that sequential procedure is reported in \cite{georgel2008unified}. However, it aims to estimate the pose parameters, which requires a calibrated camera. The objective of this paper is to estimate the projective homography, i.e., there is no calibrated camera. Furthermore, that existing technique applies a first-order minimization method, and the considered scaling factors do not take into account the convergence properties of the individual approaches, as will be proposed in the sequel.

\section{Proposed Unified Approach}
\label{sec:method}
Consider that a \emph{reference template} has been specified to an estimation algorithm. This is typically a region of interest with predefined resolution inside a larger reference image. Then, a second image, referred to as the \emph{current image}, is given to that algorithm. The goal is to find the transformation parameters that, when applied to the current image, results in a current template identical to the reference template. 

\subsection{Transformation Models}
The considered transformation models consist of a geometric and a photometric one. The geometric transformation model explains image changes due to variations in the scene structure and/or the camera motion. For a given pixel $\vect{p}^*$ in the reference template that corresponds to pixel $\vect{p}$ in the current image, we model the geometric motion using a homography:
\begin{align}
\mathbf{p} \, &\propto \, \mathbf{H} \, \mathbf{p}^*\\
&= \left[ \,
      \frac{h_{11} u^* + h_{12} v^* + h_{13}} {h_{31} u^* + h_{32} v^*
        + h_{33}}, \, \frac{h_{21} u^* + h_{22} v^* + h_{23}} {h_{31}
        u^* + h_{32} v^* + h_{33}}, \, 1 \, \right]^\top\\
&= \vect{w}(\vect{H},\vect{p}^*),
\label{equ:homog}
\end{align}
where $\mathbf{p}^* = [u^*, v^*, 1]^\top \in \mathbb{P}^2$ is the homogeneous pixel coordinates in the reference template, $\vect{w}$ is the warping operator, and $\mathbf{H} \in \mathbb{SL}(3)$ is the projective homography matrix with its elements 
$\{ h_{ij} \}$. Such matrix has only eight degrees-of-freedom. In general, this situation leads to a reprojection step after each iteration of the minimization algorithm that takes the estimated homography into the Special Linear Group. To avoid this problem, the proposed algorithm parameterizes the homography using its corresponding Lie Algebra \cite{benhimane_07}. This is accomplished via the matrix exponential function, which maps a region around the identity matrix $\vect{I} \in \mathbb{SL}(3)$ to a region around the origin $\vect{0} \in \mathfrak{sl}(3)$. A matrix $\vect{A}(\vect{v}) \in \mathfrak{sl}(3)$ is the linear combination of eight matrices that form a base of the Lie Algebra. Therefore $\vect{v}$ has eight components. A homography is thus parameterized as
\begin{equation}
    \vect{H}(\vect{v}) = \exp(\vect{A}(\vect{v})) . 
    \label{equ:matrix_exp}
\end{equation}
The homography matrix may be used to extract relative motion and scene structure information \cite{malis2007deeper}. However, this decomposition is out of the scope of this work and is unnecessary for many robotic applications.

The photometric transformation model explains the changes in the image due
to variations in the lighting conditions of the scene. Let us model in this work only global illumination variations, i.e., changes that apply
equally to all pixels in the images. This model is defined as
\begin{equation}\label{eq:I}
\mathcal{I}'(\mathbf{p}) = \alpha \, \mathcal{I}(\mathbf{p}) + \beta,
\end{equation}
where $\mathcal{I}(\mathbf{p}) \geq 0$ is the intensity value of the pixel
$\mathbf{p}$, $\mathcal{I}' (\mathbf{p}) \geq 0$ denotes its transformed
intensity, and the gain $\alpha \in \mathbb{R}$ and the bias $\beta \in \mathbb{R}$ are the
parameters that fully define the transformation. These parameters
can be viewed as the adjustments in the image constrast and brightness, respectively.

\subsection{Nonlinear Least Squares Formulation}
Consider that the reference template is composed of $m$ pixels. Also, consider that a feature detection and matching algorithm provides $n$ feature correspondences between the reference template and the current image. Ideally, it would be possible to find a vector $\vect{x}^* = \{ \vect{H}^*, \alpha^*, \beta^* \}$ such that:
\begin{align}
    \alpha^*\image(\vect{w}(\vect{H}^*,\vect{p}^*_i))+\beta^* &= \image^*(\vect{p}^*_i), \quad &\forall i &= 1, 2, \dots, m, \\
    \vect{w}(\vect{H}^*,\vect{q}^*_j) &=\vect{q}_j, \quad &\forall j &= 1, 2, \dots, n,
\end{align}
by substituting \eqref{equ:homog} in \eqref{eq:I}, where $\image$ and $\image^*$ are the current and reference images, respectively, $\vect{p}^*_i \in \mathbb{P}^2$ contains the coordinates of the $i$-th pixel of the reference template, and $\vect{q}_j, \vect{q}_j^* \in \mathbb{P}^2$ are the representations of the $j$-th feature correspondence set in the current image and reference template, respectively. The perfect calculation of $\vect{x}^*$ is impossible due to a variety of reasons, including noise in the camera sensor and outliers in the feature matching. This leads to the reformulation of this task as a nonlinear least-squares problem.

Two separate cost-functions are defined: One for the IB part and another for the FB one. The $i$-th pixel of the reference template contributes to the following row to the IB cost function via the distance
\begin{equation}
    a_i(\vect{x}) = \alpha\image(\vect{w}(\vect{H},\vect{p}^*_i))+\beta - \image^*(\vect{p}^*_i),
\end{equation}
and an output vector $\vect{y}_{IB}$ can be constructed as:
\begin{equation}
    \vect{y}_{IB} = \begin{bmatrix} a_1 &  a_2 & \cdots & a_m
     \end{bmatrix}^\top.
              \label{equ:cost_ib}
\end{equation}
The FB cost function is defined using the distance between the features coordinates in each image:
\begin{equation}
     \vect{b}_j(\vect{x}) = \vect{w}(\vect{H},\vect{q}^*_j) -\vect{q}_j = \begin{bmatrix} b^u_j & b^v_j & 0
     \end{bmatrix},
\end{equation}
where $b^u_j, b^v_j$ are distances between the features in the $u$ and $v$ directions, respectively. The third element is disregarded since it is always zero. Thus, a vector $\vect{y}_{FB}$ can be constructed as:
\begin{equation}
    \vect{y}_{FB} = \begin{bmatrix} b^u_1 &  b^v_1&  b^u_2 &  b^v_2 & \cdots & b^u_n & b^v_n
     \end{bmatrix}^\top.
         \label{equ:cost_fb}
\end{equation}

Using \eqref{equ:cost_ib} and \eqref{equ:cost_fb}, a unified nonlinear least squares problem can be defined as
\begin{equation}
     \min_{\vect{x} = \{\vect{H}, \alpha, \beta\}}  \ \frac{1}{2} 
     \Bigl( 
      w_{IB}\norm{ \vect{y}_{IB}(\vect{x}) }^2_2 + 
      w_{FB}\norm{ \vect{y}_{FB}(\vect{x}) }^2_2
      \Bigr),
     \label{eq:min_initial_unified}
\end{equation}
where $w_{IB}, w_{FB}$ are carefully chosen weights given to the intensity- and feature-based components of the cost function, respectively, as will be proposed later on. For real-time systems, only local optimization methods can be applied since global ones are too costly. In this case, an initial approximation $\widehat{\vect{x}} = \{ \widehat{\vect{H}}, \widehat{\alpha}, \widehat{\beta} \}$ of the true solution is required. This estimate can be integrated into the least-squares formulation as:
\begin{equation}
     \min_{\vect{z} = \{\vect{v}, \alpha, \beta\}}  \ \frac{1}{2} 
     \Bigl( 
      w_{IB}\norm{ \vect{y}_{IB}(\vect{x}(\vect{z}) \circ \widehat{\vect{x}}) }^2 + 
      w_{FB}\norm{ \vect{y}_{FB}(\vect{x}(\vect{z}) \circ \widehat{\vect{x}}) }^2
      \Bigr),
     \label{eq:min_incremental_unified}
\end{equation}
where the symbol `$\circ$' denotes the composition operation. For the scalars $\alpha$ and $\beta$, it corresponds to the addition, whereas for the homography that operation is the matrix multiplication. Furthermore, to take into account the different number of observations for IB and FB methods, we include normalization factors and define the unified output vector as
\begin{equation}
    \vect{y}_{UN} =
    \begin{bmatrix}
     \sqrt{\frac{w_{IB}}{m}} \vect{y}_{IB} &
     \sqrt{\frac{w_{FB}}{2n}} \vect{y}_{FB}
    \end{bmatrix}.
\end{equation}
Hence, a more concise unified formulation is achieved:
\begin{equation}
     \min_{\vect{z} = \{\vect{v}, \alpha, \beta\}}  \ \frac{1}{2} 
      \norm{ \vect{y}_{UN}(\vect{x}(\vect{z}) \circ \widehat{\vect{x}})}^2,
     \label{eq:min_incremental_unified_concise}
\end{equation}
which can be efficiently solved using \cite{silveira_ijcv10}.

\subsection{Weight Choices}

The weights $w_{IB}$ and $w_{FB}$ should be carefully selected to ensure the best convergence properties for the algorithm. The following constraints apply to the weights:
\begin{align}\label{eq:convex}
    w_{IB} + w_{FB} &= 1,\\ w_{FB}, w_{IB} &> 0.
\end{align}
The idea behind the proposed method for determining the weights is to let the feature-based error be more influential to the optimization when the current solution is far from the true one. As the FB error decreases, then the intensity-based component becomes increasingly more important. This is consistent with the idea that the FB method is better suited to handle large displacements, whereas IB methods have higher accuracy, but only work when the initial guess is sufficiently close to the true solution.

The main measurement used for calculating the weights is the feature-based error associated with the current estimated homography $\widehat{\vect{H}}$. It is calculated using the following root mean squared error (RMSD):
\begin{equation}
    RMSD(\vect{y}_{FB}) = \sqrt{ \frac{\sum^n_{j=1} \norm{\vect{w}(\widehat{\vect{H}},\vect{q}^*_j) -\vect{q}_j}^2_2}{n}} = d_{FB}.
\end{equation}
The proposed weights are then defined from
\begin{equation}
    w_{FB} = 1 - \exp(-d_{FB})
\end{equation}
and \eqref{eq:convex}. This function allows for a continuous transition where the feature-based weight decreases as its error gets lower, and the intensity-based component becomes increasingly more important in the optimization.

\subsection{Local versus Global Search}
\label{sec:global}

The processing times may be drastically increased if the feature detection and matching algorithms are allowed to process the entire current image. The proposed method processes only a small region in the current image to obtain good matches whenever possible.

Firstly, a current template is generated by warping the current image with the initial approximation $\widehat{\vect{H}}$. Then, this current template is assigned a score by comparing it with the reference template using the Zero-mean Normalized Cross-Correlated. If this score is higher than a predefined threshold, then the feature detection algorithm searches only within this current template. Otherwise, the current template and $\widehat{\vect{H}}$ are both discarded. In this case, the detection algorithm searches the entire current image for features. The first scenario is referred to as a ``local'' search, whereas the second one as a ``global'' search. When the global search is used, it is necessary to recalculate an initial approximation $\widehat{\vect{H}}$. This is done by calculating the homography solely from the features matches between the current image and the reference template.

\section{Experimental Results}
\label{sec:results}
\subsection{Validation Setup}

\begin{figure}[!ht]
  \centering
  \copyrightbox[b]{\includegraphics[width=0.9\linewidth]{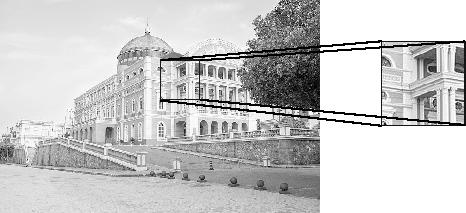}}{``Teatro Amazonas Atualmente 01'' by Karine Hermes | Modified}\\[2mm]
  \includegraphics[width=.49\linewidth]{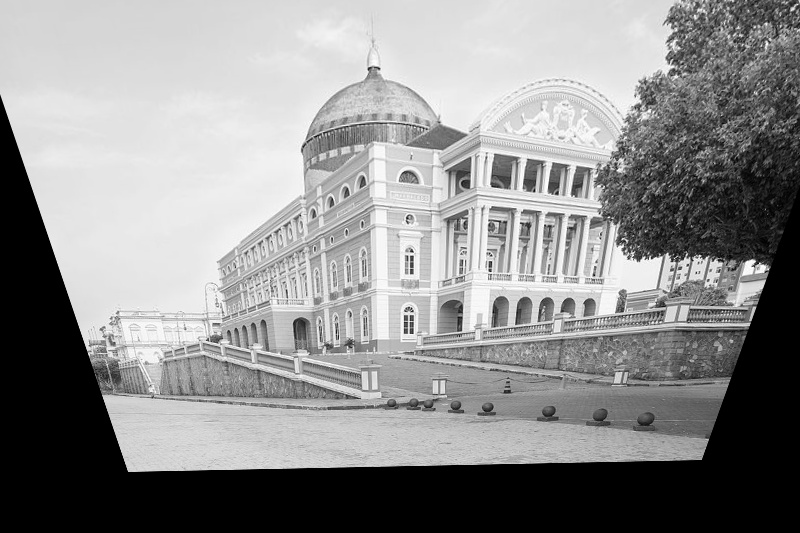}  
  \includegraphics[width=.49\linewidth]{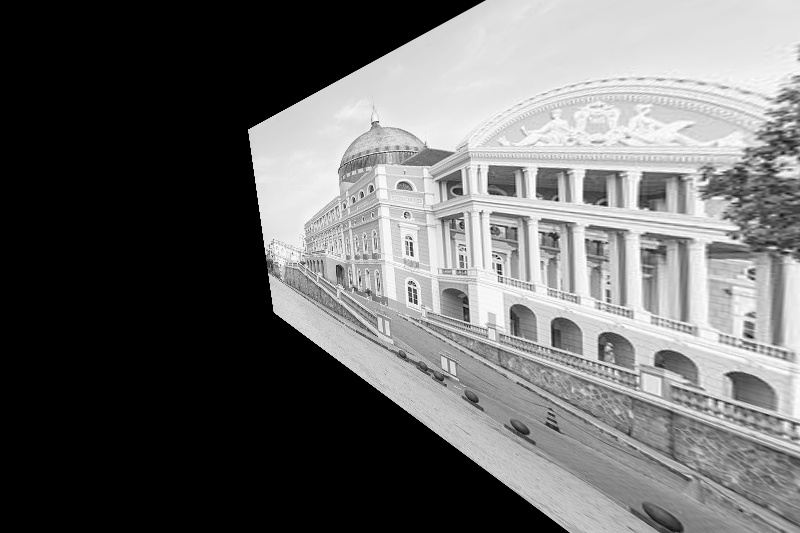}  
\caption{Validation setup. (Top) Reference image and selected reference template, resp. (Bottom) Examples of transformation with perturbations $\sigma = 5$ and $\sigma = 10$, resp.}
\label{fig:validation_setup}
\end{figure}

The same testing procedure used in \cite{baker2001equivalence} is implemented to validate the algorithm. Firstly, a reference image of size $800 \times 533$ pixels is chosen, and a region of size $100 \times 100$ pixels is selected as the reference template. The coordinates of each corner are independently perturbed in the $\overrightarrow{u}$ and $\overrightarrow{v}$ directions with a zero mean Gaussian noise and standard deviation of $\sigma$ pixels (see Figure~\ref{fig:validation_setup}). The relation between the original corner points and the perturbed ones defines a test homography. The reference image is then transformed by this test homography. The algorithm receives the reference template and the transformed image with the identity element as the initial guess for the photogeometric transformation. From this input, the algorithm produces an estimated homography. In turn, this homography is used to transform each reference corner point. If the average residual error between the actual perturbed corner points and the estimated perturbed ones is less than 1 pixel, the result is declared to have converged. 1,000 test cases are randomly generated for each value of the perturbation $\sigma \in [0, 20]$ and used as input for each evaluated algorithm. In all tests, 3 levels of a multiresolution pyramid are used. In each level, a maximum of 3 iterations of the algorithm are allowed to execute.

This setup is used to compare different algorithms. Three criteria are analyzed: Convergence domain, convergence rate and timing analysis. The methods differ on whether they use only the IB or the FB component (SURF is here applied for feature detection and description) in the cost function, or both for the Unified case. Another difference is the use of a ZNCC predictor to improve the initialization in some methods. Finally, some algorithms do not consider the photometric part of the transformation space. These algorithms along with their characteristics are summarized in Table~\ref{table:methods}. 

\medskip

\begin{table}[!ht]
\captionsetup{font={footnotesize,sc},justification=centering,labelsep=period}
\caption{Homography estimation algorithms used for comparisons.}
\centering%
\small
\begin{tabular}{lcccc}
     Method & IB & FB & Predictor & Photometric \\ \midrule
     ESM &  \cmark & \xmark  & \xmark & \xmark \\
     \myrowcolour
     IBG  &  \cmark & \xmark  & \xmark& \cmark\\
     IBG\_P & \cmark & \xmark  & \cmark& \cmark\\
     \myrowcolour
     FB\_ESM & \xmark & \cmark  & \xmark& \xmark\\
     UNIF& \cmark & \cmark  & \xmark& \cmark\\
     \myrowcolour
     UNIF\_P& \cmark & \cmark  & \cmark& \cmark
\end{tabular}
\label{table:methods}   
\end{table}

\subsection{Convergence Domain}
\label{sec:dominio}

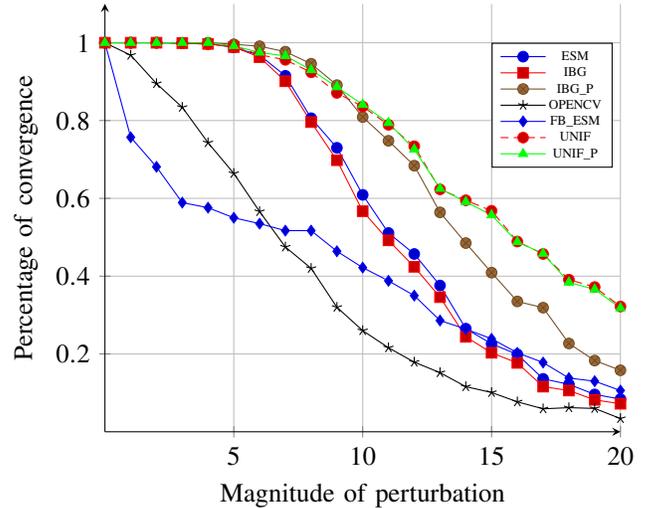
\begin{figure}[!ht]
\begin{center}
\begin{tikzpicture}[thick, scale=1.0]
\begin{axis}[
axis lines=middle,
ymin=0,
ymax=1.1,
xmax=20.0,
grid=both,
ylabel near ticks,
xlabel near ticks,
xlabel=Magnitude of perturbation,
ylabel= Percentage of convergence,
enlargelimits = false,
legend style={at={(axis cs:15.0,1.0)},anchor=north west, nodes={scale=0.5, transform shape}},]
\addplot table [y=ib_basic,x=Sigma]{data/convergence_all.dat};
\addlegendentry{ESM}
\addplot table [y=ibg_regular,x=Sigma]{data/convergence_all.dat};
\addlegendentry{IBG}
\addplot table [y=ibg_predictor,x=Sigma]{data/convergence_all.dat};
\addlegendentry{IBG\_P}
\addplot table [y=fb_cv,x=Sigma]{data/convergence_all.dat};
\addlegendentry{OPENCV}
 \addplot table [y=robust_fb,x=Sigma]{data/convergence_all.dat};
\addlegendentry{FB\_ESM}
\addplot table [y=unified,x=Sigma]{data/convergence_all.dat};
\addlegendentry{UNIF}
\addplot[color=green, mark=triangle*] table [y=unified_predictor,x=Sigma]{data/convergence_all.dat};
\addlegendentry{UNIF\_P}
\end{axis}
\end{tikzpicture}
\caption{Percentage of convergence versus magnitude of perturbation for different homography estimation algorithms.}
\label{fig:convergence_all}
\end{center}
\end{figure}

Figure~\ref{fig:convergence_all} shows that the proposed Unified algorithms have a larger convergence domain than all pure FB or IB versions. It also shows that the use of the ZNCC predictor in the unified version does not affect its frequence of convergence, as well as that the IBG (i.e., IB with robustness to Global illumination changes) and ESM algorithms have a very similar performance. The latter is expected because there are no lighting changes in this validation setup.

Another interesting observation is that the results of the algorithms in the FB class (FB\_ESM and the algorithm available in OpenCV) were significantly worse than the ones in the IB class, although it was expected that they would have a higher convergence domain. This suggests that there is still room for improving the FB components of the estimation, which would in turn lead to a further improvement in the unified method as well.

\subsection{Convergence Rate}
\label{sec:rate}

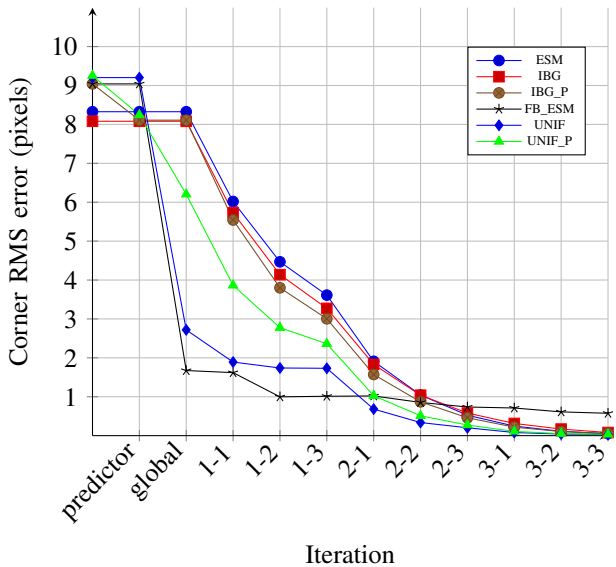
\begin{figure}[!ht]
\begin{center}
\begin{tikzpicture}[thick, scale=1.0]
\begin{axis}[
axis lines=middle,
ymin=0,
ymax=11.0,
xmin=0,
xmax=11.0,
grid=both,
ylabel near ticks,
xlabel near ticks,
xlabel= Iteration,
ylabel= Corner RMS error (pixels),
enlargelimits = false,
xtick=data,
xticklabels={0, predictor, global, 1-1, 1-2, 1-3, 2-1, 2-2, 2-3, 3-1, 3-2, 3-3},
ytick={1,2,3,4,5,6,7,8,9,10},
x tick label style={
            rotate=45,
            anchor=east,
        },
legend style={at={(axis cs:8.0,10.0)},anchor=north west, nodes={scale=0.5,transform shape}},]
\addplot table [y=ib_basic,x=Iter]{data/rate_all.dat};
\addlegendentry{ESM}
\addplot table [y=ibg_regular,x=Iter]{data/rate_all.dat};
\addlegendentry{IBG}
\addplot table [y=ibg_predictor,x=Iter]{data/rate_all.dat};
\addlegendentry{IBG\_P}
\addplot table [y=robust_fb,x=Iter]{data/rate_all.dat};
\addlegendentry{FB\_ESM}
\addplot table [y=unified,x=Iter]{data/rate_all.dat};
\addlegendentry{UNIF}
\addplot[color=green, mark=triangle*] table [y=unified_predictor,x=Iter]{data/rate_all.dat};
\addlegendentry{UNIF\_P}
\end{axis}
\end{tikzpicture}
\caption{Pixel RMS error after each optimization iteration for different homography estimation algorithms under perturbation $\sigma=10$.}
\label{fig:rate_all}
\end{center}
\end{figure}

Figure~\ref{fig:rate_all} compares the convergence rate of the homography estimation algorithms under a perturbation of magnitude $\sigma=10$. This rate is displayed as the progression of the root mean squared (RMS) error between the coordinates of the 4 corners of the reference template and the estimated transformation of the current template. Out of the 1,000 test cases, only those where the estimation converged are considered here. Note that the results from the OpenCV algorithm is omitted because it was used as a black-box, and therefore the sequence of
homographies at each iteration cannot be accessed. The $x$-axis of Figure~\ref{fig:rate_all} contains each important step in the optimization. The first step, which is labeled ``predictor'', is the result of the ZNCC prediction step. The second step, which is labeled ``global'', is the step where the algorithm decides to search for features in the entire current image, as described in Section~\ref{sec:global}. Of course, these two steps are not performed by every algorithm. Afterwards, steps from the iterative optimization method follow. They are separated by pyramids level, such that the notation ``X-Y'' represents pyramid level X at iteration Y.

Figure~\ref{fig:rate_all} allows for several observations. Firstly, the FB\_ESM performance is very dependent on the ``global'' step. After this step, it is the algorithm with the best RMS value. However, it is not capable to improve this value too much in the subsequent optimization steps. When the other algorithms reach the third level of the pyramid, they all outperform its RMS. The behaviour of ESM, IBG and IBG\_P is very similar as  they share the same framework. A small difference between them is that IBG\_P is able to converge even for cases with a slightly higher initial RMS error, due to the prediction step. After that step, however, all these three algorithms perform quite similarly.

Finally, let us note that the Unified algorithms have a behaviour that combines the FB and IB methods, as desired. The UNIF\_P uses both the ``predictor'' and ``global'' steps. Interestingly, the global search is less applied in that version than the UNIF one because of the prediction step. This explains its smaller initial reduction in RMS value. On the other hand, less usage of the global step leads to a improvement in the processing times, as shown in the next section. After these steps, both the Unified algorithms behave similarly to IB ones, with the advantage of having a better initialization procedure.

\subsection{Timing analysis}
\label{sec:timing}

\begin{figure}[!ht]
\begin{center}
\begin{tikzpicture}[thick, scale=1.0]
\begin{axis}[
axis lines=middle,
ymin=0,
ymax=0.16,
xmax=20.0,
grid=both,
ylabel near ticks,
xlabel near ticks,
xlabel= Magnitude of Perturbation,
ylabel= Processing Time (s),
enlargelimits = false,
ytick={0,0.02, 0.04, 0.06, 0.08, 0.1, 0.12, 0.14, 0.16},
yticklabel style={
        /pgf/number format/fixed,
        /pgf/number format/precision=2
},
legend style={at={(axis cs:15.0,0.08)},anchor=north west, nodes={scale=0.5, transform shape}},]
\addplot table [y=ib_basic,x=Sigma]{data/timing_all.dat};
\addlegendentry{ESM}
\addplot table [y=ibg_regular,x=Sigma]{data/timing_all.dat};
\addlegendentry{IBG}
\addplot table [y=ibg_predictor,x=Sigma]{data/timing_all.dat};
\addlegendentry{IBG\_P}
\addplot table [y=robust_fb,x=Sigma]{data/timing_all.dat};
\addlegendentry{FB\_ESM}
\addplot table [y=unified,x=Sigma]{data/timing_all.dat};
\addlegendentry{UNIF}
\addplot[color=green, mark=triangle*] table [y=unified_predictor,x=Sigma]{data/timing_all.dat};
\addlegendentry{UNIF\_P}
\end{axis}
\end{tikzpicture}
\caption{Processing times for different perturbation levels.}
\label{fig:timing_all}
\end{center}
\end{figure}
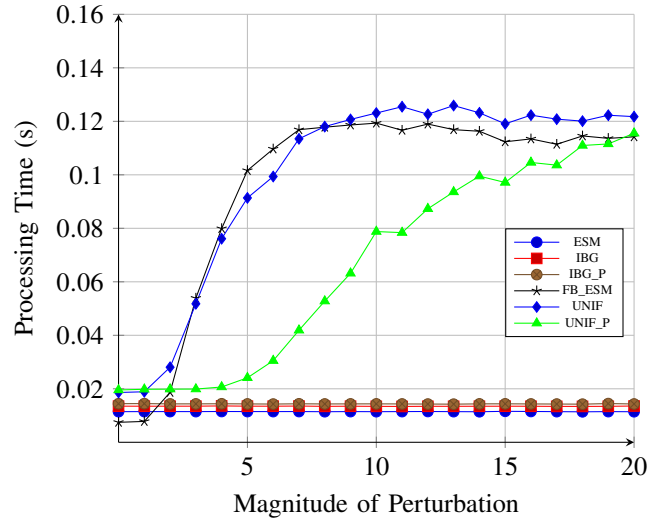

Figure~\ref{fig:timing_all} shows how the average time needed to run the estimation algorithms varies depending on the magnitude of perturbation. This time is measured in a Intel i7-6700HQ processor, and is averaged over the subset of the 1,000 cases only when the estimation has converged. The most noticeable aspect of this graph is that pure IB algorithms have nearly constant time, regardless of the perturbation level. In constrast, the algorithms that have a feature-based component need more time to process images with higher perturbation levels. This phenomenon can be explained by considering the effect of the global versus local feature search. As the perturbation level increases, the number of occasions where the algorithm applies the global search also increases. This step, however, is very computationally expensive. The UNIF\_P manages to have a lower processing time because the prediction step increases the probability that the local search is used. Therefore, the UNIF\_P can be seen as a compromise between having the advantage of being capable of performing global search, without taking a big penalty in the processing times.

However, these results also show that more research is needed to develop a method that is able to reliably perform in real-time settings for large perturbations. The IB methods are already capable of that when they converge, requiring less than 0.02s/image. The FB and Unified methods may need up to 0.12s, which may be unacceptable for some applications.

\begin{figure*}[!ht]
\centering
\includegraphics[width=0.16\linewidth]{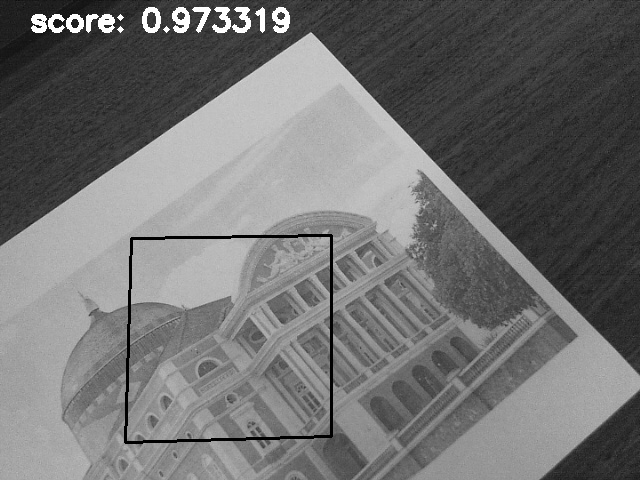}
\includegraphics[width=0.16\linewidth]{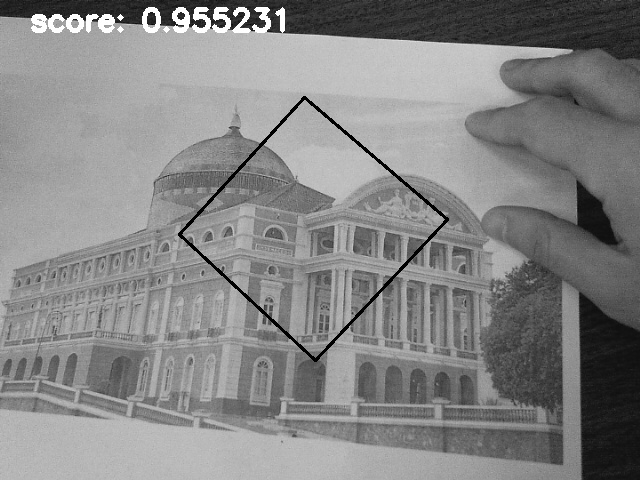}
\includegraphics[width=0.16\linewidth]{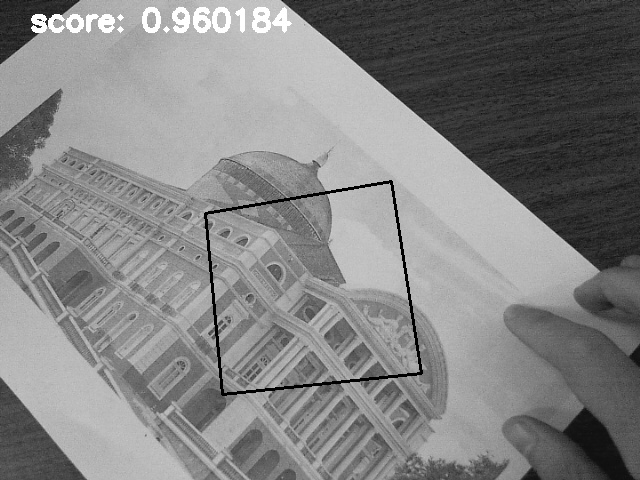}
\includegraphics[width=0.16\linewidth]{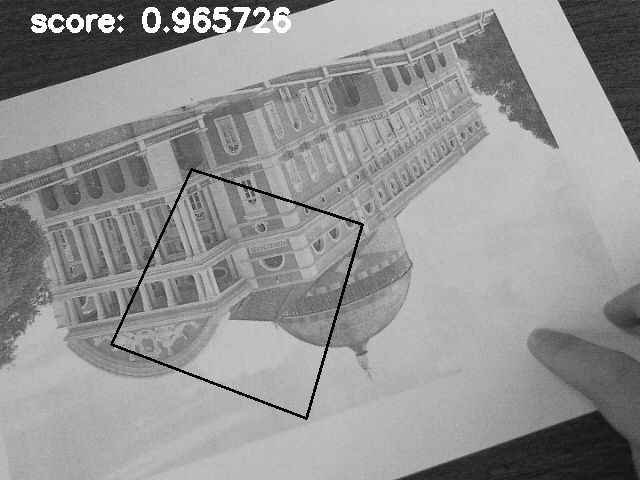}
\includegraphics[width=0.16\linewidth]{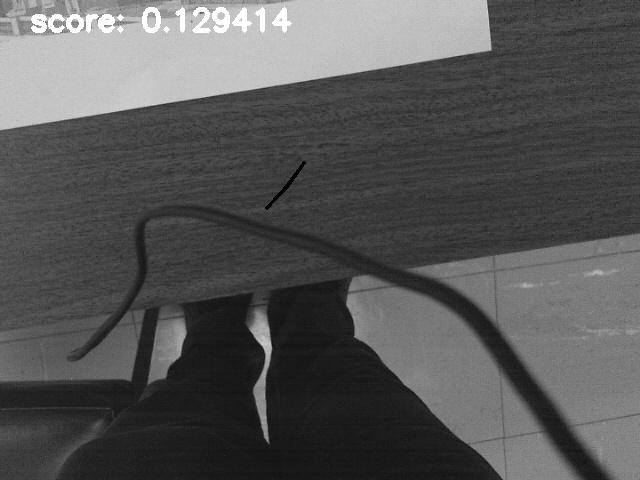}
\includegraphics[width=0.16\linewidth]{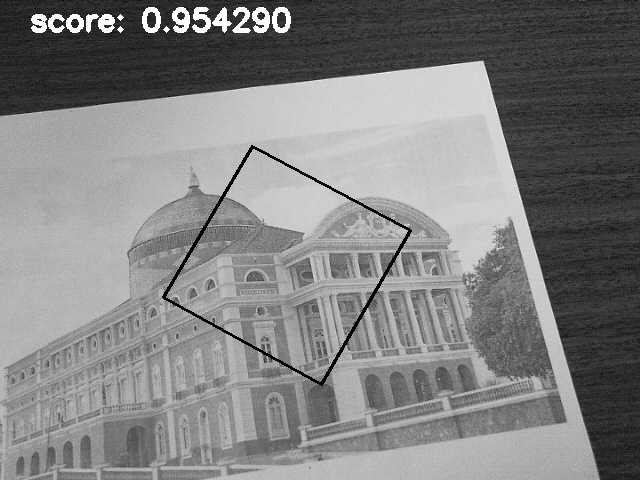}\\[1mm]
\includegraphics[width=0.16\linewidth]{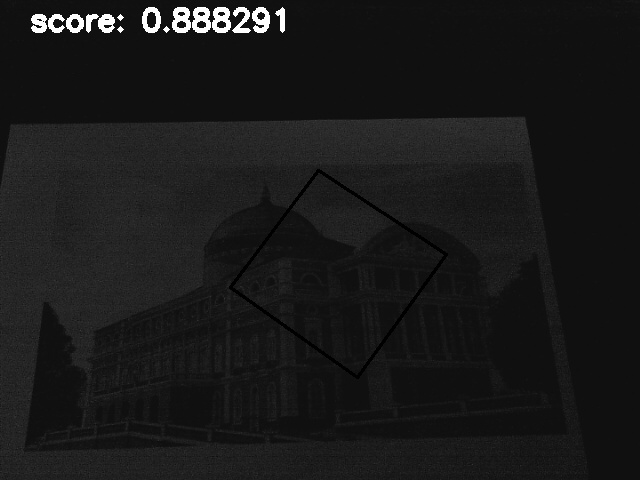}
\includegraphics[width=0.16\linewidth]{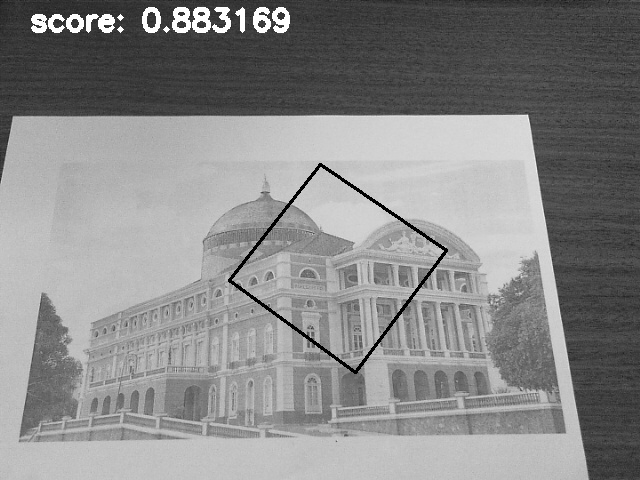}
\includegraphics[width=0.16\linewidth]{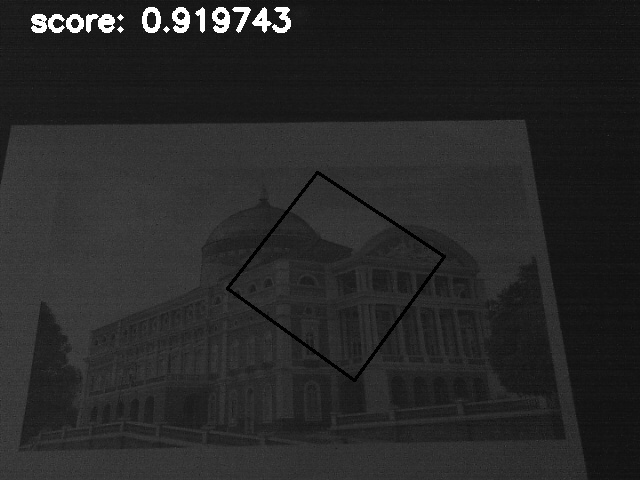}
\includegraphics[width=0.16\linewidth]{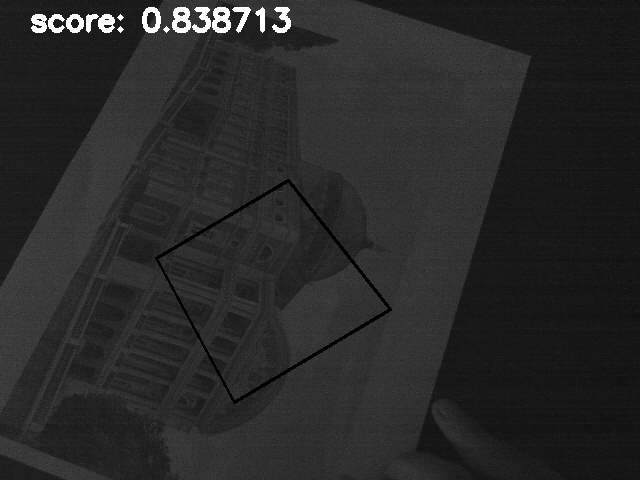}
\includegraphics[width=0.16\linewidth]{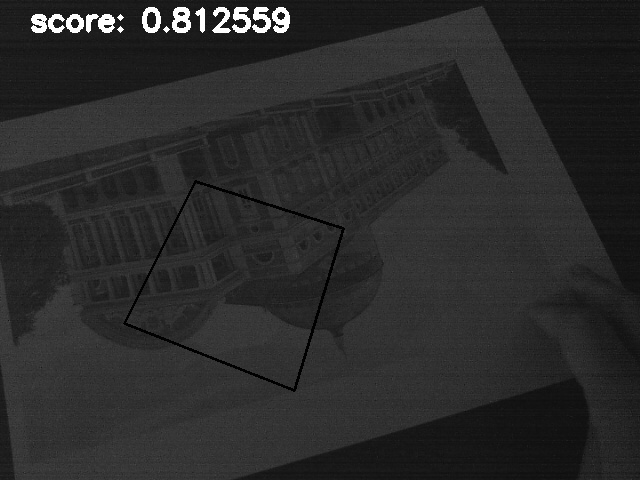}
\includegraphics[width=0.16\linewidth]{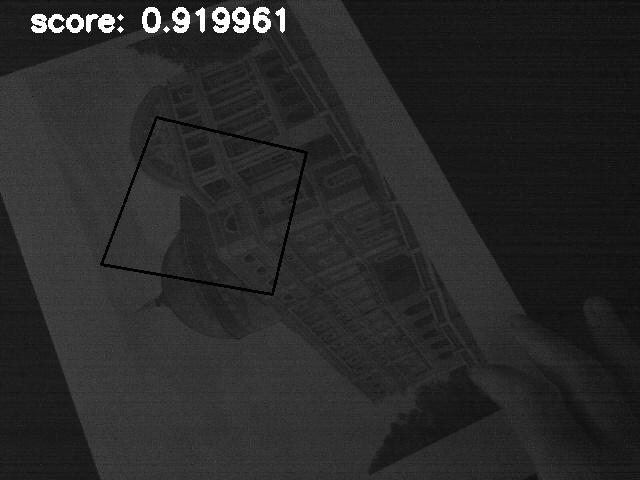}\\[1mm]
\includegraphics[width=0.16\linewidth]{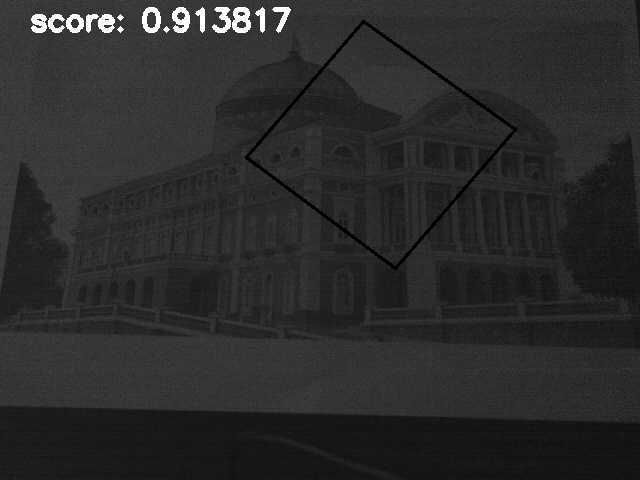}
\includegraphics[width=0.16\linewidth]{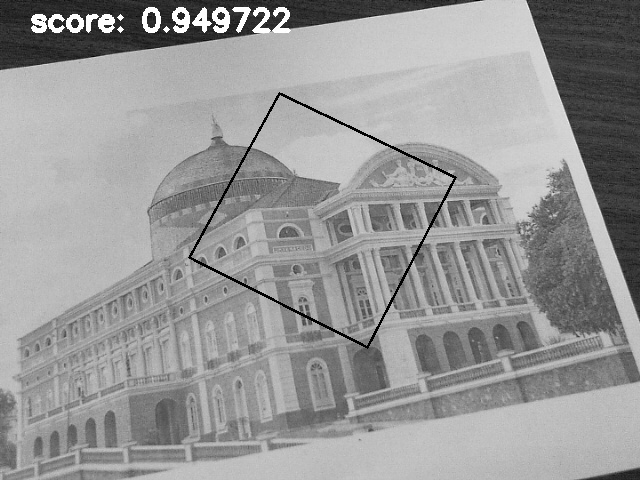}
\includegraphics[width=0.16\linewidth]{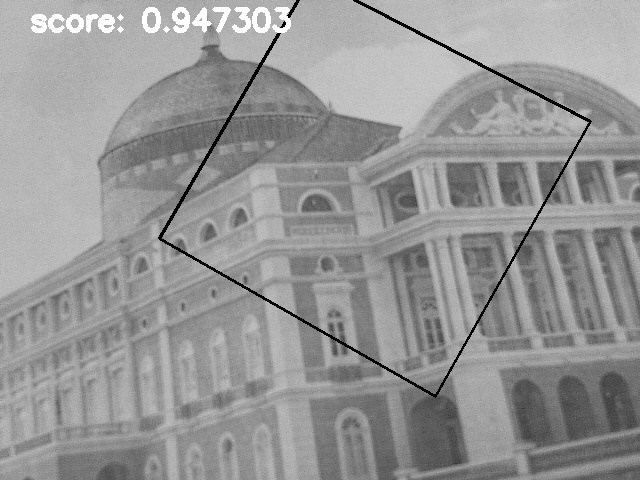}
\includegraphics[width=0.16\linewidth]{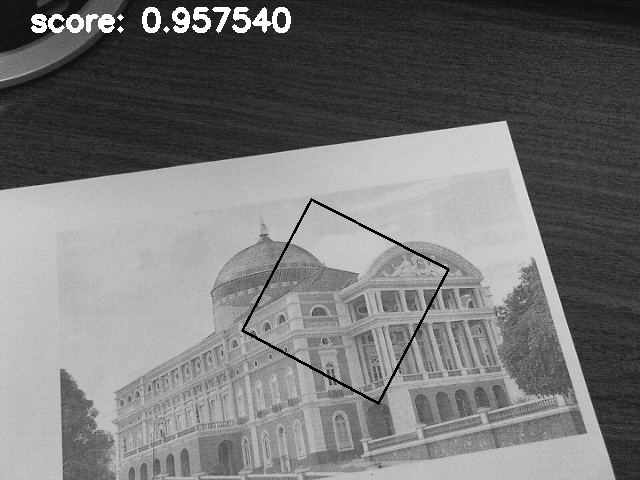}
\includegraphics[width=0.16\linewidth]{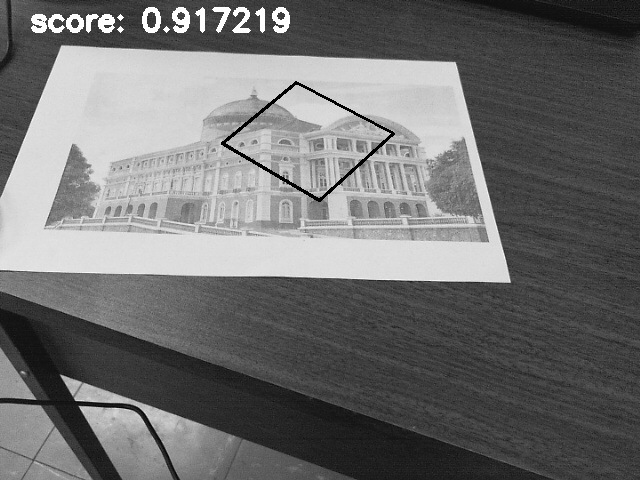}
\includegraphics[width=0.16\linewidth]{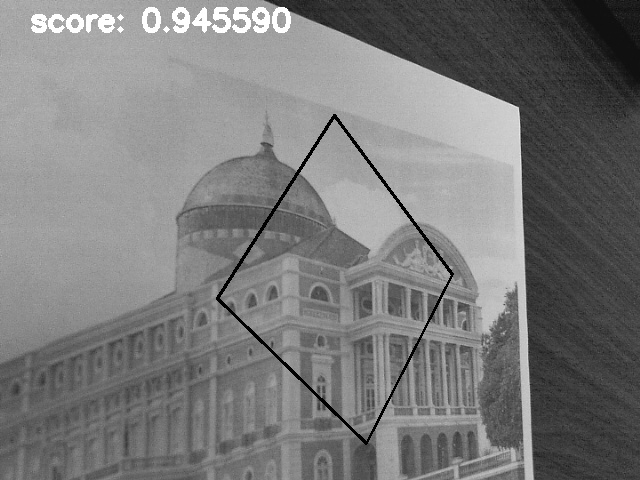}
\caption{Excerpts of homography-based visual tracking (left-to-right then top-to-bottom) using the proposed unified approach.}
\label{fig:unified_tracking}
\end{figure*}

\subsection{Use Case: Visual Tracking}
\label{sec:usecase}
The proposed algorithm is publicly available for research purposes as a C++ library and as a ROS package \cite{Webpage:2020:VTEC_ROS}, along with its technical report \cite{nogueira2019}. This section shows its application to homography-based visual tracking. Results are available at \cite{Webpage:2020:youtube_video}. The prediction step is applied, as recommended for real-time tracking applications. Figure~\ref{fig:unified_tracking} shows some excerpts of this tracking experiment. An interesting result is that the proposed unified visual tracker can recover from full occlusions. Even after completely removing the tracked region from the current image, the tracker can recover given its feature-based ability to perform the ``global'' search. Additionally, it can be seen that the algorithm is robust to large global illumination changes, and that in some cases it can recover from complete failure even under severe lighting variations.

\section{Conclusions}\label{sec:conc}

This paper proposes a first step towards a truly unified optimal approach to homography estimation. The results show that improved convergence properties are indeed obtained when combining both classes of feature- and intensity-based methods into a single optimization procedure. This can help vision-based applications to handle faster robot motions. Future work will focus on reducing the processing time of the unified algorithm, specially when very large interframe displacements lead to a global search for features.

\section*{Acknowledgment}

This work was supported in part by the CAPES under Grant 88887.136349/2017-00, in part by the FAPESP under Grant 2017/22603-0, and in part by the InSAC project (CNPq under Grant 465755/2014-3, FAPESP under Grant 2014/50851-0).



%
%
%

\bibliographystyle{IEEEtran}
\bibliography{bareconf.bib}


\end{document}